\documentclass[letterpaper, 10 pt, conference]{ieeeconf}  

\IEEEoverridecommandlockouts   

\usepackage{authblk}
\usepackage{graphics} 
\usepackage{epsfig} 
\usepackage{mathptmx} 
\usepackage[font=small]{caption}
\usepackage{times} 
\usepackage{amsmath} 
\usepackage{amssymb}  
\usepackage{cite}
\usepackage{xcolor}
\usepackage{booktabs}
\usepackage{float}
\usepackage{multirow}
\usepackage{stfloats}
\usepackage{caption}
\usepackage{newtxtext, newtxmath}
\usepackage{tabularx}
\usepackage{comment}
\usepackage{longtable}
\usepackage[pagebackref=true,breaklinks=true,colorlinks,citecolor=gray]{hyperref}

\usepackage{flushend}
\usepackage{algorithm}
\usepackage{algpseudocode}
\usepackage{subcaption}
\usepackage{pifont}
\newcommand{\cmark}{\ding{51}} 
\newcommand{\xmark}{\ding{55}} 

\title{\LARGE \bf
Flexible Locomotion Learning with Diffusion Model Predictive Control
}

\author{Runhan Huang$^{\ddagger}$, Haldun Balim$^{\ddagger}$, Heng Yang$^{\ddagger}$, Yilun Du$^{\ddagger\dagger}$\\
\href{https://Flexible-Diffusion-MPC.github.io/}{\large\textbf{Flexible-Diffusion-MPC.github.io/}}
\thanks{$^{\ddagger}$Kempner Institute, Harvard University, MA, USA}%
\thanks{$^{\dagger}$ Corresponding at: \texttt{ydu@seas.harvard.edu}}%
}

\begin{document}

\newcommand{\insertteaser}{
    \centering
    \includegraphics[width=\linewidth]{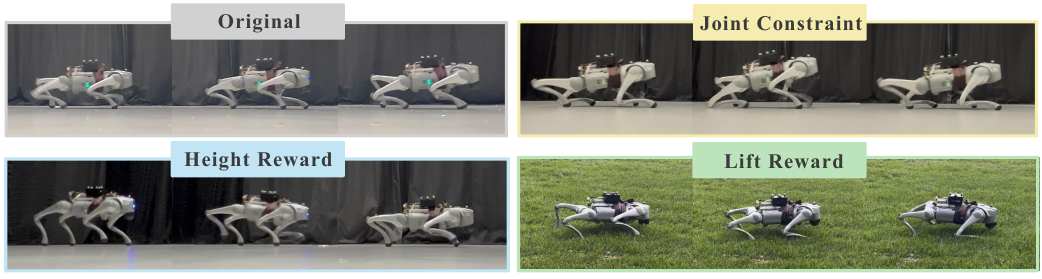}
    \vspace{-15pt}
    \captionof{figure}{\textbf{Flexible Adaptation of Locomotion Policies.} Our learned model predictive control procedure allows the quadruped to flexibly adapt its behavior at test time to new joint constraints, heights, and terrains.}
    \label{fig:teaser}
    \vspace{-15pt}
}

\makeatletter
\apptocmd{\@maketitle}{\centering\insertteaser}{}{}
\makeatother

\maketitle
\setcounter{figure}{1}

\thispagestyle{empty}
\pagestyle{empty}

\begin{abstract}
Legged locomotion demands controllers that are both robust and adaptable, while remaining compatible with task and safety considerations. However, model-free reinforcement learning (RL) methods often yield a \emph{fixed} policy that can be difficult to adapt to new behaviors at test time. In contrast, Model Predictive Control (MPC) provides a natural approach to flexible behavior synthesis by incorporating different objectives and constraints directly into its optimization process. However, classical MPC relies on accurate dynamics models, which are often difficult to obtain in complex environments and typically require simplifying assumptions. We present \emph{Diffusion-MPC}, which leverages a learned generative diffusion model as an approximate dynamics prior for planning, enabling flexible test-time adaptation through reward and constraint based optimization. Diffusion-MPC jointly predicts future states and actions; at each reverse step, we incorporate reward planning and impose constraint projection, yielding trajectories that satisfy task objectives while remaining within physical limits. To obtain a planning model that adapts beyond imitation pretraining, we introduce an \emph{interactive training} algorithm for diffusion based planner: we execute our reward-and-constraint planner in environment, then \emph{filter and reweight} the collected trajectories by their realized returns before updating the denoiser. Our design enables strong test-time adaptability, allowing the planner to adjust to new reward specifications without retraining. We validate Diffusion-MPC on real world, demonstrating strong locomotion and flexible adaptation.
\end{abstract}

\section{INTRODUCTION}
\label{sec:intro}
Legged locomotion remains a significant challenge in robotics, as controllers must guarantee stability under complex contact-rich dynamics~\cite{kumar2021rma, lee2020learning} while accommodating evolving task requirements. A generalized controller demands the synthesis of diverse behaviors—ranging from posture adjustment~\cite{margolis2023walk, ai2025towards} and balancing~\cite{zhang2025hub} to energy-efficient walking~\cite{nai2025fine}. This notion of flexible behavior synthesis~\cite{janner2022planning} is central to enabling robots to operate autonomously in unpredictable environments. Rather than being limited to a single, pre-defined policy, a capable locomotion controller should be able to integrate new constraints, and adapt to novel objectives on the fly~\cite{janner2022planning}. Such flexibility is especially critical at test time, where robots may face new terrains, altered physical limits, or modified task goals, which can not be anticipated during training or even design phase. Meeting this demand requires not only robustness to disturbances but also rapid test-time adaptation—the ability to reshape trajectories in real time.

Building toward this vision, the field has converged on two main solutions to locomotion control. Model predictive control \cite{rawlings2017model} offers a natural mechanism for adaptation by explicitly encoding objectives and constraints into an optimization-based control framework. However, MPC depends on highly accurate dynamics models, incurs significant computational cost in contact-rich settings, and in practice often requires extensive simplifications and hand-crafted design choices \cite{bledt2018cheetah, neunert2018whole}. By contrast, model-free reinforcement learning~\cite{schulman2017proximal, haarnoja2018soft} has achieved impressive results for behaviors~\cite{kumar2021rma, zhuang2023robot, margolis2023walk}, producing robust policies that can transfer from simulation to hardware; yet these methods typically learn a fixed policy network, making it difficult to flexibly synthesize multiple behaviors at test time and thus most successful in relatively static tasks. Together, these limitations highlight the gap between RL’s efficiency and MPC’s flexibility, motivating approaches that combine their respective strengths.

We address this gap by introducing \textbf{Diffusion-MPC}, which leverages generative diffusion models~\cite{ho2020denoising, janner2022planning, song2019generative, song2020denoising} as expressive learned priors, enabling them to function as planners that implicitly capture system dynamics. Rather than directly fitting an action only policy, our diffusion model learns to jointly represent state transitions and action proposals from large, heterogeneous datasets. This learned generative prior then plays the role of the planner in an MPC framework: during each planning cycle, trajectories are sampled from the diffusion model and optimized with reward terms and constraints, effectively performing model-based planning without reliance on hand-crafted dynamics. In this view, diffusion models are not just conditional generators, but expressive approximators of environment dynamics that make tractable, flexible MPC possible. Reward-based planning updates steer generated trajectories toward task objectives, while feasibility is maintained through constraint projection. Candidate ranking is then applied to further refine the selected plan. Together, these mechanisms provide adaptability while avoiding the need for simplified model designs required by MPC and the rigidity of fixed RL policies. In addition, the framework naturally supports skill compositionality: multiple reward and constraint terms can be combined at test time to synthesize behaviors without retraining.

Directly training a diffusion planner on demonstrations, however, often fails to meet deployment requirements. In practice, datasets not only may lack coverage of critical objectives such as energy efficiency or smoothness, but may also suffer from limited quality or inconsistencies, since there is rarely access to a true expert capable of providing optimal demonstrations~\cite{cao2021learning}. Furthermore, demonstrations are typically collected in simulation under simplified physics, whereas deployment occurs in the real world where conditions such as terrain inclination, contact friction, and unmodeled disturbances differ substantially~\cite{kumar2021rma}. As a result, purely imitative rollouts may drift off-distribution during real~\cite{huang2024diffuseloco}. To address this, we introduce an interactive online training procedure for diffusion based planners: the diffusion planner is rolled out interactively to collect trajectories, and the model is updated using trajectories filtered and reweighted by realized returns, in the spirit of reward-weighted regression~\cite{peters2007reinforcement, kang2023efficient, peng2019advantage, ren2024diffusion}. This adaptation both enhances robustness and improves alignment with task objectives, while preserving the compositional interface for flexible test-time control.

In summary, to tackle the challenge of achieving flexible and adaptable locomotion, we propose a novel framework that makes the following key contributions:

\begin{itemize}
\item \textbf{Diffusion-MPC Formulation.} We reinterpret diffusion models as generative priors for model predictive control via reward-based updates, constraint projection, and candidate ranking.
\item \textbf{Interactive Training.} We propose a reward-weighted denoising adaptation procedure that finetunes the diffusion planner during interaction with the environment, enhancing locomotion capability.
\item \textbf{Real-World Deployment.} We design practical techniques for real-time diffusion planning, including asynchronous execution and early-step caching, and demonstrate zero-shot transfer on a Unitree Go2 quadruped.
\end{itemize}

\section{RELATED WORKS}
\subsection{Learning-Based Locomotion}
\looseness=-1
Two main paradigms dominate locomotion control: model predictive control and reinforcement learning. MPC approaches for legged locomotion~\cite{bledt2018cheetah,neunert2018whole} optimize trajectories with explicit costs and constraints, enabling agile and versatile maneuvers. However, these methods typically rely on simplified dynamics models, which limits accuracy in contact-rich locomotion and makes them computationally demanding, often necessitating substantial modeling simplifications. By contrast, model-free RL~\cite{schulman2017proximal, haarnoja2018soft} has achieved impressive results through massively parallel simulation training~\cite{makoviychuk2021isaac, gu2023maniskill2, tao2024maniskill3}. The learned policies was successfully deployed in different scenarios, including traversing complex terrains~\cite{kumar2021rma, wu2023learning, zhu2025robust, cheng2024quadruped, ren2025vb}, achieving extreme motions~\cite{he2024agile, zhang2025hub, huang2025moe, zhuang2023robot, luo2024pie, cheng2024extreme, hoeller2024anymal}, and interactively navigate in the real world~\cite{jin2023resilient, zhu2025vr, zhu2025saro, uppal2024spin}. Yet policies learned via model-free RL are tightly coupled to their training reward functions, confining them to predefined behaviors and hindering adaptation to novel test-time objectives. Hence the generalizability and adaptability of RL is not desirable for flexible behavior synthesis. Works such as~\cite{margolis2023walk} further shows that reinforcement learning can endow a single policy with a family of behaviors and interpolate between them. However, the set of behaviors must be predefined through task-specific reward design and integrate into network input during training, limiting flexibility when new objectives arise at deployment. Building on the powerful expressiveness of generative models~\cite{ho2020denoising, song2020denoising}, recent approaches~\cite{chi2023diffusion, janner2022planning} have explored their use as planners that capture system dynamics implicitly. In this view, diffusion models provide a learned generative prior over trajectories, which can then be steered by reward terms and constraints at test time to enable flexible adaptation.

\subsection{Diffusion for Control}
Diffusion models~\cite{sohl2015deep,ho2020denoising, bansal2023universal, black2023training, peebles2023scalable} have recently been explored as generative frameworks for decision making~\cite{chi2023diffusion, janner2022planning, ajay2022conditional, qi2025strengthening, romer2024diffusion, chen2024diffusion, luo2025generative, karunratanakul2023guided, huang2025diffuse, luo2024potential}. Recently, there have been approaches to leverage the expressiveness of diffusion in locomotion control. ~\cite{huang2024diffuseloco} use diffusion policy as action generator to learn from large corpus of demonstrations. \cite{o2025offline} adopt a classifier free guidance with return on the diffusion policy to control the robot toward different behavior, but still have to predefine one-hot skill vectors as network input and incorporate pre-defined return in training time, which limits the adaptability and flexibility of behavior. ~\cite{zhou2025diffusion} has shown that diffusion models can serve as generative priors for planning, but the framework focused on relatively simple planning and was evaluated in simulation. Compared to the action-only diffusion motion generation~\cite{huang2024diffuseloco, o2025offline}, we incorporate implicit dynamics modeling through modeling state-action distribution and conduct planning during deployment, enabling flexible behavior at test-time.

\section{PROBLEM SETUP AND BACKGROUND}
We consider a discrete-time dynamical system
\[
s_{t+1} = f(s_t, a_t),
\]
where $s_t \in \mathbb{R}^{n_s}$ denotes the state and $a_t \in \mathbb{R}^{n_a}$ the control input at time $t$.  
The objective is to maximize a cumulative reward $r(s_t,a_t)$ while ensuring that the trajectories satisfy constraints $(s_t,a_t) \in \mathcal{C}$.  

Since optimizing over long horizons is typically computationally prohibitive, MPC~\cite{rawlings2017model} instead considers a finite and tractable horizon of length $H$. At each time step, MPC plans over this horizon by solving
\begin{subequations}
    \begin{align*}
        \max_{a_{t:t+H-1 \mid t}} \;\; &\sum_{i=0}^{H-1} r(s_{t+i\mid t}, a_{t+i\mid t})
    \end{align*}
\end{subequations}
subject to the dynamics constraint $s_{t+i+1\mid t} = f(s_{t+i\mid t}, a_{t+i\mid t})$ and the constraint $(s_{t+i\mid t}, a_{t+i\mid t}) \in \mathcal{C}$ for $i = 0,\dots,H-1$, starting from the current state $s_t$. After solving this optimization, only the first action $a_{t\mid t}$ is applied, and the problem is re-solved at the next step in a receding-horizon fashion.

While classical MPC frameworks rely on solving an optimization problem at each step, a recently popular alternative is to cast planning as a conditional generation problem~\cite{janner2022planning,ajay2022conditional}. Concretely, we consider that we have access to a trajectory dataset  $\mathcal{D} = \{ \xi_i \}_{i=1}^N, \quad \xi_i = \{(s_t^i, a_t^i)\}_{t=0}^T,$
collected from diverse policies that are not necessarily optimal nor guaranteed to satisfy the constraints in $\mathcal{C}$. From this dataset, we learn a generative model that samples state–action sequences of horizon $H$:  
\[
\tau =
\begin{bmatrix}
s_{t} & s_{t+1} & \cdots & s_{t+H} \\
a_{t} & a_{t+1} & \cdots & a_{t+H}
\end{bmatrix},
\quad
\tau \in \mathbb{R}^{(n_s + n_a) \times H},
\]
aimed at maximizing rewards while respecting constraints. Unlike optimization-based MPC, this sampling-based approach learns a generative prior over trajectories, allowing rewards and constraints to be incorporated directly during sampling.

\textbf{Trajectory diffusion.}
Diffusion models~\cite{sohl2015deep} provide a flexible generative framework by gradually perturbing data with noise and learning a reverse denoising process to recover structured samples. Beyond image and signal domains, they have also been successfully applied to trajectory modeling and control~\cite{janner2022planning, ajay2022conditional}. Motivated by these successes, we employ diffusion models to learn a generative prior over trajectories. Specifically, we consider a data-generating process:
\begin{equation*}
q\!\left(\tau^{(k)} \mid \tau^{(k-1)}\right)
= \mathcal{N}\!\Big(\sqrt{\alpha_k}\,\tau^{(k-1)},\; (1-\alpha_k)\,\mathbf{I}\Big),
\end{equation*}
and a learned reverse process:
\begin{equation*}
p_\theta\!\left(\tau^{(k-1)} \mid \tau^{(k)}\right)
= \mathcal{N}\!\Big(\mu_\theta(\tau^{(k)},k),\;\Sigma_k\Big),
\end{equation*}
with a pre-defined variance schedule $\{\alpha\}_{k=0}^K \in (0,1]$ and a neural network $\mu_\theta$. To learn the parameters $\theta$, we use the denoising objective~\cite{ho2020denoising}:  
\begin{equation*}
    \mathcal{L}(\theta) = \mathbb{E}_{\tau, k, \epsilon}\,\big\| \tau - \tau_\theta(\tau^{(k)}, k) \big\|^2,
\end{equation*}
where $\tau^{(k)}$ is obtained from the forward noising process and $\epsilon \sim \mathcal{N}(0, I)$.  

\vspace{-3pt}
\section{METHOD}

\subsection{Flexible Behavior Synthesis Through Sampling.}

In this section, we present our methodology for synthesizing reward-maximizing, constraint-satisfying trajectories using our learned generative prior. Given a differentiable trajectory-level reward function $R(\tau)$, we define the distribution of desired trajectories as  
\begin{equation*}
\pi(\tau) \propto p_\theta(\tau)\,\exp\!\big(\lambda R(\tau)\big)\,\mathbf{1}\{\tau \in \mathcal{C}\},
\end{equation*}
where $p_\theta(\tau)$ is the learned generative prior by our diffusion model, $\lambda \geq 0$ controls the strength of planning, and $\mathbf{1}\{\tau \in \mathcal{C}\}$ enforces constraints.  

To sample from $\pi$, we tilt the diffusion prior $p_\theta(\tau)$ using reward-based planning, following the reward planning approach of~\cite{janner2022planning}. After each reverse step, we apply a reward planning update:  
\begin{equation*}
 \tau^{(k-1)} \coloneq \tau^{(k-1)} + \eta_k\,\Sigma_k\,\nabla_{\tau} R\!\big(\tau^{(k-1)}\big),
\end{equation*}
where $\Sigma_k$ is the covariance of the reverse transition and $\eta_k$ controls the strength of planning.  

For the trajectory-level reward $R(\tau)$, we combine two complementary signals: a learned neural component $R_{\mathrm{nn}}$ that captures long-horizon semantic behaviors difficult to formalize analytically, and an analytic component $R_{\mathrm{an}}$ that encodes hand-specified objectives.  
\[
R(\tau) = \alpha_{\mathrm{nn}} R_{\mathrm{nn}}(\tau) + \alpha_{\mathrm{an}} R_{\mathrm{an}}(\tau).
\]
This combination offers flexibility: neural rewards provide expressive, data-driven semantics, while analytic rewards enable direct incorporation of task-specific structure. Together, they allow planning to balance learned behavior with explicitly defined objectives.

To ensure constraint, we project the denoised sample onto the feasible set after each step in reverse chain:  
\begin{equation*}
\tau^{(k-1)} \coloneq \Pi_{\mathcal{C}}\!\left(\tau^{(k-1)}\right),
\end{equation*}
where $\Pi_{\mathcal{C}}$ denotes the projection operator onto $\mathcal{C}$. This mechanism provides a flexible way to enforce constraint, ensuring that sampled trajectories remain valid while still allowing diverse behaviors to emerge.     

\looseness=-1
The overall sampling procedure is shown in Algorithm~\ref{alg:diffusion_mpc}. 
Starting from Gaussian noise, the reverse process iteratively refines a candidate trajectory. 
At each step, we incorporate reward functions to plan the trajectory distribution toward high-reward behaviors and enforce constraint through projection. 
The initial state is fixed throughout the process to ensure consistency with the current observation. 
This combination of denoising and planning yields state–action trajectories that both adhere to constraints and flexibly adapt to task objectives.  


\begin{figure}[h]
\vspace{-15pt}
\begin{minipage}{\linewidth}
\begin{algorithm}[H]
\caption{Diffusion-MPC Sampling}
\label{alg:diffusion_mpc}
\begin{algorithmic}[1]
\Require diffusion model $\mu_\theta$, reward planning scale $\lambda$, reward function $R(\tau)$, constraint set $\mathcal{C}$, initial state $s_0$, number of candidates $N$
\State \textit{All updates below are applied in parallel to the $N$ candidates.}
\State initialize $\tau^{K}\!\sim\!\mathcal{N}(0,I)$ with batch size $N$; set first state $\tau^{K}_{s_0}\!\leftarrow\! s_0$
\For{$k=K,\dots,1$} 
    \State ${\tau}^{k-1}\!\leftarrow\!\mu_\theta(\tau^{k},k)+\sigma_k \epsilon_k,\;\; \epsilon_k\!\sim\!\mathcal{N}(0,I)$ \Comment{reverse step}
    \State $\tau^{k-1}\!\leftarrow\!{\tau}^{k-1}+\lambda\,\Sigma_k\,\nabla_{\tau}R(\tau^{k-1})$ \Comment{reward planning}
    \State $\tau^{k-1}\!\leftarrow\!\Pi_{\mathcal{C}}\!\big({\tau}^{k-1}\big)$ \Comment{constraint projection}
    \State $\tau^{k-1}_{s_0} \leftarrow s_0$ \Comment{enforce initial state}
\EndFor
\State $\tau^\star = \arg\max_{\tau\in\{\tau_i^0\}} R(\tau)$\Comment{ranking across N candidates}
\end{algorithmic}
\end{algorithm}
\end{minipage}
\vspace{-5pt}
\end{figure}


In addition to reward-based planning and constraint injection, we employ a candidate ranking strategy: multiple trajectories are sampled from the diffusion planning process, scored by the reward, and the best one selected. This complements local gradient updates by improving robustness through exploration of diverse rollouts. This global selection complements local gradient planning: while gradient steps provide fine-grained planning, candidate ranking improves robustness by exploring diverse rollouts and mitigating noisy gradients planning.

Compared with action-only diffusion approaches~\cite{chi2023diffusion} that rely on separately conditioned dynamics models~\cite{zhou2025diffusion}, our joint state–action formulation provides a unified representation of trajectories and controls. By generating states and actions together, the diffusion prior naturally respects system dynamics, eliminating the need for an explicit dynamics model. Moreover, this formulation enables reward shaping and constraint injection to act directly on the evolving state sequence at every denoising step, rather than only through actions. This joint perspective increases flexibility in incorporating task objectives and feasibility constraints.

\subsection{Compositional Behavior Synthesis}
Beyond planning with a single reward function, our framework also supports flexible skill composition through reward combination.  
Let $\{R_i(\tau)\}_{i=1}^K$ denote a set of scalar task rewards, which may be either neural or analytic.  
At deployment, the user specifies weights $\alpha \in \mathbb{R}^K$ to form a composite objective
\begin{equation*}
R_{\alpha}(\tau) = \sum_{i=1}^{K} \alpha_i\,R_i(\tau).
\end{equation*}
By varying the weights $\alpha_i$, Diffusion MPC can seamlessly trade off between different objectives, synthesizing a diverse range of behaviors.  
This includes not only behaviors represented in the dataset but also novel behaviors arising from new combinations of reward signals.

\subsection{Planner Learning with Environment Interaction} \label{sec:finetune}

We propose a strategy to collect data and finetune the diffusion prior using trajectories generated by our model in an online interactive way.
Let $(\tau^{(k)}, k, \epsilon)$ denote a standard denoising tuple constructed from a clean trajectory $\tau$ with forward noise $\epsilon \sim \mathcal{N}(0,I)$.
Per-trajectory weights are defined from realized returns as
\begin{equation*}
w\!\left(R_r(\tau)\right)=\exp\!\left(\frac{R_r(\tau)}{T}\right),
\end{equation*}
with $T>0$ as a temperature parameter. Here $R_r$ denotes the ground-truth return from the environment, analogous to the RL setting, rather than the reward model used during planning. To filter out low-return rollouts, we retain only the top-$K$ weights and set the rest to zero:
\begin{equation*}
w'(\tau) =
\begin{cases}
w(R_r(\tau)), & \tau \in \text{Top-}K, \\
0, & \text{otherwise.}
\end{cases}
\end{equation*}
\begin{equation*}
\bar w(\tau) = \frac{w'(\tau)}{\mathbb{E}[w'(\tau)]}.
\end{equation*}
The resulting objective is
\begin{equation*}
\mathcal{L}_{\text{RWD}}(\theta)=\mathbb{E}\Big[\,\bar w\!\left(R_r(\tau)\right)\,\big\|\,\tau-\tau_{\theta}(\tau^{(k)},k)\big\|_2^2\Big].
\end{equation*}
which performs exponentially tilted regression, biasing updates toward higher-return trajectories.
A replay buffer is maintained that interleaves on-policy planner rollouts with previously collected trajectories, preserving the coverage of prior experiences while nudging the model toward reward-favored regions of the trajectory space.

Compared with standard policy-gradient methods, reward-weighted denoising has several advantages for diffusion planners: (i) it remains in the native denoising parameterization, avoiding high-variance policy gradients and critic bootstrapping; (ii) it is naturally off-policy and sample-efficient—filter and reweight can exploit stored trajectories without importance-correction instabilities; (iii) it is flexible to explicitly plan into several behavior in the sampling phase, giving strong guidance during the rollout. In practice, this yields a stable online procedure that trains a planner flexibly while preserving the benefits of the learned generative prior.

\vspace{-3pt}
\subsection{Real-time Planning}

\textbf{Asynchronous planning for real-time control.}
To meet high-rate locomotion requirements, we employ an asynchronous pipeline with planning horizon \(H\) and replan margin \(D\).
At each timestep \(t\), the controller executes the next action from the current \(H\)-step plan \(a_{t:t+H-1}\).
When the execution index reaches \(H{-}D\), we trigger replanning from the latest observation to synthesize a fresh \(H\)-step plan while continuing to execute the remaining \(D\) buffered actions from the old plan.
Once these \(D\) actions have been applied, we time-align the new plan by skipping its first \(D\) actions and begin execution at offset \(D\).
Equivalently, each action is computed \(D\) control cycles before it is applied (a \(D\)-step action buffer), which maintains real-time operation while preserving closed-loop feedback with period \(H{-}D\) steps.
In our experiments, we set \(H{=}11\) and \(D{=}3\).

\textbf{Caching early denoising.} Successive plans generated by our model often produce nearly identical trajectories in early diffusion steps, as these steps primarily denoise without incorporating task-specific structure. To avoid redundant computation, we shift the existing plan across time steps and reuse it as the initialization for the next window, up to $m$ steps. This warm-start strategy preserves solution quality while substantially reducing inference cost.  

\textbf{Sampler choice and step budget.}  
DDIM offers faster, deterministic sampling at some cost in fidelity, while DDPM is slower but higher quality. The number of denoising steps controls the compute–quality trade-off. We use $10$ DDPM steps at inference and ablate both the training horizon and test-time step count.

\vspace{-3pt}
\section{EXPERIMENTS: DESIGN AND SETUP}

\subsection{Experiment Setup}
For adaptation tasks, we consider locomotion tasks with objectives: base height variation, joint limit restriction, energy saving, joint acceleration/velocity regularization, and balancing. We use Isaacgym\cite{makoviychuk2021isaac} as our simulator for pretrain dataset collection and interactive training. The planners are initialized with dataset collected with domain randomization. Our interactive diffusion planner training in \ref{subsec:online_learning_result} is conducted on a single NVIDIA 3090 GPU, with 4096 environments in simulation. The control frequency in both the simulation environment and the real world is 50Hz. Our policy is deployed on a Unitree Go2 quadrupedal robot, with an Intel NUC 12 PRO as onboard computing device. We use PD control for low-level joint torques $(K_p=40.0, K_d=1.0)$. The robot observation at time $t$ is represented as
\begin{equation*}
o_t = \{\, v^{\text{yaw}}_t,\; g_t,\; v^{\text{cmd}}_t,\; q_t,\; \dot{q}_t,\; a_{t-1} \,\},
\end{equation*}
where $v^{\text{yaw}}_t$ is the base angular velocity, $g_t$ is the projected gravity,
$v^{\text{cmd}}_t$ is the command vector, $q_t$ and $\dot{q}_t$ are the joint position and velocity,
and $a_{t-1}$ is the previous action.

\vspace{-5pt}
\subsection{Adaptation Task Definition}

\subsubsection{Base Height Variation}
Base height is not included in the observation. We therefore use a trajectory-level reward model
\begin{equation*}
R_{\text{height}}(\tau;\,h^\star) \;=\; f_{\phi}\!\big(\tau^{(k)},\,t\big)\,,
\end{equation*}
where a U-Net takes a corrupted joint state–action segment $\tau^{(k)}$ and diffusion step $t$ and predicts the reward of the corresponding clean trajectory; at test time, $h^\star$ denotes the desired height profile implicit in the query. In our setting, we set $h^{\star} = 0.15$ m.

\subsubsection{Joint Limit Restriction}
We consider (i) attraction to a target posture $q^{\text{tar}}$ via an analytic term
\begin{equation*}
R_{\text{posture}}(\tau) \;=\; -\frac{1}{H}\sum_{t=0}^{H-1}\|q_t-q^{\text{tar}}\|_2^2,
\end{equation*}
and (ii) a tightened range $q^{\min}\!\le q_t\!\le q^{\max}$ enforced by reward
\begin{equation*}
R_{\text{range}}(\tau) = -\frac{1}{H}\sum_{t=0}^{H-1}
\Bigl(
  \bigl\|[\,q_t - q^{\max}\,]_+\bigr\|_2^2
 +\bigl\|[\,q^{\min} - q_t\,]_+\bigr\|_2^2
\Bigr)
\end{equation*}
and also projection
\begin{equation*}
q_t \leftarrow \Pi_{[\,q^{\min},\,q^{\max}\,]}(q_t).
\end{equation*}
where $[u]_+=\max(u,0)$ and $\Pi_{[a,b]}(q)=\min(\max(q,a),b)$ acts elementwise.

\subsubsection{Energy Saving}
We train a reward model $R_{\text{energy}}(\tau)=f_{\psi}(\tau)$ to predict energy cost and plan accordingly. The reward is calculated by the time-integrated mechanical power
\begin{equation*}
R_{\text{energy}}(\tau)\;=\;-\sum_{t=0}^{H-1}\sum_{j=1}^{d_u}\big|\tau_{j,t}\,\dot{q}_{j,t}\big|\,\Delta t,
\end{equation*}
with joint torque $\tau_{j,t}$ and velocity $\dot{q}_{j,t}$.

\subsubsection{Joint Acceleration / Velocity Regularization}
We penalize high rates with analytic terms
\begin{equation*}
R_{\text{vel/acc}}(\tau)\;=\;-\lambda_v\sum_{t=0}^{H-1}\|\dot{q}_t\|_2^2\;-\;\lambda_a\sum_{t=1}^{H-1}\left\|\frac{\dot{q}_t-\dot{q}_{t-1}}{\Delta t}\right\|_2^2,
\end{equation*}
and enforce rate limits by projection $\dot{q}_t\!\leftarrow\!\Pi_{[-\dot{q}^{\max},\,\dot{q}^{\max}]}(\dot{q}_t)$.

\subsubsection{Balancing}
We align the gravity direction measured in the body frame with a desired unit direction. Let $g_b(s_t)$ be the gravity vector expressed in the body frame (available from IMU/state), and define the unit vector $\hat g_t = g_b(s_t)/\|g_b(s_t)\|_2$. Let $\hat d_t$ be the desired unit gravity direction (for level posture, $\hat d_t = [0,0,-1]^\top$). We use
\begin{align*}
R_{\text{align}}(\tau)  &= -\frac{1}{H}\sum_{t=0}^{H-1}\big(1 - \hat g_t^\top \hat d_t\big),\\
R_{\text{smooth}}(\tau) &= -\lambda_{\text{tv}}\sum_{t=1}^{H-1}\|\hat g_t - \hat g_{t-1}\|_1,\\
R_{\text{balance}}(\tau) &= R_{\text{align}}(\tau) + R_{\text{smooth}}(\tau).
\end{align*}

\vspace{-5pt}
\subsection{Learning the Planner}
\looseness=-1
Our training follows a two-stage procedure. In the first stage, we pretrain the planner on $4{,}000$ trajectories of length $1{,}000$, collected from a demonstrator policy trained with PPO. The demonstration data is collected in a static environment which is not sufficiently representative for robust deployment. 
To address this, we finetune the planner as described in Sec.~\ref{sec:finetune}, using $1500$ additional environment interactions for each environments. During this stage, we apply domain randomization with the parameters detailed in Tab.~\ref{tab:random}.

\begin{table}[htbp]
    \renewcommand{\arraystretch}{1.0}
    \fontsize{8}{9.5}\selectfont
    \centering
    \begin{tabular}{c|cc}
        \toprule
        \textbf{Parameters} & \textbf{Range} & \textbf{Unit} \\ \midrule
        Base mass & [1, 3] & $kg$\\
        Mass position of X axis & [-0.2, 0.2] & $m$\\
        Mass position of Y axis & [-0.1, 0.1] & $m$\\
        Mass position of Z axis & [-0.05, 0.05] & $m$\\
        Friction & [0, 2] & - \\
        Initial joint positions & [0.5, 1.5]$~\times~$nominal value & $rad$\\
        Motor strength & [0.9, 1.1]$~\times~$nominal value & - \\
        Proprioception latency & [0.005, 0.045] & $s$\\
        \bottomrule
    \end{tabular}
    \vspace{-1mm}
    \caption{\label{tab:random}\textbf{Domain Randomization} The parameters used for domain randomization during finetuning phase.}
    \vspace{-10pt}
\end{table}

\section{EXPERIMENTS: RESULTS}
\subsection{Simulation Experiments}
\subsubsection{Adaptation Tasks}

\begin{table*}[htbp]
    \vspace{2mm}
    \centering
    \setlength\tabcolsep{3pt} 
    \fontsize{9}{11.5}\selectfont
    \begin{tabular}{l|cccccccc}
        \toprule
        \multirow{2}{*}{\textbf{Objectives}} & \multicolumn{7}{c}{\textbf{Penalty} $\downarrow$} \\
        & \textbf{Diffusion Policy} 
        & \textbf{C1 R\xmark ~C\xmark} 
        & \textbf{C10 R\xmark ~C\xmark} 
        & \textbf{C100 R\xmark ~C\xmark} 
        & \textbf{C1 R\cmark ~C\xmark } 
        & \textbf{C10 R\cmark ~C\xmark} 
        & \textbf{C100 R\cmark ~C\xmark} 
        & \textbf{C1 R\xmark ~C\cmark} \\
        \midrule
        \textbf{Joint Vel} 
        & 0.969 & 0.973 & 0.752 & 0.743 
        & 0.762 & 0.697 & 0.684 & \textbf{0.553} \\
        \midrule
        \textbf{Joint Acc} 
        & 0.350 & 0.349 & 0.263 & 0.254 
        & 0.267 & 0.259 & \textbf{0.249} & 0.262 \\
        \midrule
        \textbf{Joint Pos (N)} 
        & 0.051 & 0.484 & 0.348 & 0.343 
        & 0.296 & 0.317 & 0.307 & \textbf{0.183} \\
        \midrule
        \textbf{Joint Pos (P)} 
        & 0.243 & 0.240 & 0.238 & 0.237 
        & 0.209 & 0.208 & 0.204 & \textbf{0.168} \\
        \midrule
        \textbf{Balancing} 
        & 0.807 & 0.768 & 0.663 & 0.650 
        & 0.562 & 0.564 & \textbf{0.551} & 0.662 \\
        \midrule
        \textbf{Energy} 
        & 0.774 & 0.825 & 0.664 & 0.626 
        & 0.519 & 0.498 & \textbf{0.482} & 0.700 \\
        \bottomrule
    \end{tabular}
    \caption{\textbf{Adaptation Performance of Diffusion-MPC Planner.} 
    Metrics report penalties for different task components, with each penalty type scaled independently for clarity. Smaller penalties indicate closer adherence to the desired behavior. Results are shown as a function of candidate number (Cand), reward-based planning (R), and constraint enforcement (C). For example, \textbf{C10 R\cmark ~C\xmark} indicates 10 candidates, reward-based planning enabled, and constraints disabled.}
    \label{tab:planner_metrics}
    \vspace{-10pt}
\end{table*}
The adaptation capability of Diffusion-MPC is assessed across a range of objectives using 1,000 environments over 3,000 simulation steps, with the average penalty reported as the evaluation metric. The study systematically varies the number of candidate trajectories together with the inclusion of reward-based planning and constraint projection to examine their individual and combined contributions. Joint Pos (N) refers to tasks with negative rear-leg joint position targets, while Joint Pos (P) corresponds to positive targets. To ensure fair assessment, both linear and angular velocity tracking are maintained within 97 percent of the baseline policy without planning, preventing excessive trade-off between task adaptation and nominal locomotion quality. Tracking performance is quantified using an exponential error metric, where the squared difference between commanded and realized velocities is penalized and mapped through $\exp(-\|e\|^2/\sigma)$, yielding values close to one for accurate tracking and approaching zero otherwise.

Table~\ref{tab:planner_metrics} demonstrates the adaptation ability of Diffusion-MPC across variations in candidate number, reward planning, and constraint planning. All values are normalized penalties, with less penalty denoting better alignment to desired behaviors. Increasing the candidate number consistently improves performance by enabling more diverse trajectory proposals. Reward planning further reduces penalties across all task categories, demonstrating that reward planning provides broadly beneficial trajectory shaping beyond candidate diversity. Constraint projection is most critical for feasibility-dominated tasks such as joint position adaptation, directly enforcing safe joint configurations. The penalty remains nonzero not due to failure, but because constraints intentionally extend beyond the demonstration distribution; in these regions, our planner successfully generalizes to discover relatively feasible actions consistent with the planning objective, though the resulting penalty is not exactly zero. Overall, candidate diversity improves robustness, reward planning drives semantic alignment, and constraints projection further guarantee feasibility; together they provide complementary benefits and confirm the design principles of Diffusion-MPC.

\subsubsection{Interactive Learning Experiments}

We compare a planner initialized on offline data and subsequently finetuned via interactive training against a baseline initialized on the same offline data without finetuning. Each configuration is evaluated in $1000$ parallel environments for $1000$ steps, and we report success rate and mean velocity deviation. As shown in Table~\ref{tab:finetune_forward}, finetuning improves stability and velocity-tracking accuracy, with the largest gains at high commanded speeds.

\begin{table}[t]
    \centering
    
    \setlength\tabcolsep{8pt}
    \fontsize{8.5}{11}\selectfont
    \begin{tabular}{l|l|cc}
        \toprule
        \textbf{Goal (Task)} & \textbf{Metric} & \textbf{finetuned} & \textbf{w/o finetuning} \\
        \midrule
        \multirow{2}{*}{\textbf{0.5 m/s Forward}} 
        & Stability (\%) & \textbf{99.7} & 46.6 \\
        & $E_v$ (\%) & \textbf{29.6} & 51.4 \\
        \midrule
        \multirow{2}{*}{\textbf{0.7 m/s Forward}} 
        & Stability (\%) & \textbf{90.5} & 23.1 \\
        & $E_v$ (\%) & \textbf{28.5} & 77.0 \\
        \midrule
        \multirow{2}{*}{\textbf{1.0 m/s Forward}} 
        & Stability (\%) & \textbf{98.8} & 18.4 \\
        & $E_v$ (\%) & \textbf{29.1} & 74.8 \\
        \bottomrule
    \end{tabular}
    \caption{\textbf{Effectiveness of Finetuning.} 
Results show stability (percentage of successful runs) and average velocity deviation ($E_v$) across different commanded forward speeds.}
    \label{tab:finetune_forward}
    \vspace{-18pt}
\end{table}

\subsection{Real-world Experiments}
\subsubsection{Adaptation Tasks}
\label{subsec:adaptation_result}
Adaptation capability is evaluated across four representative tasks: energy saving, joint position regulation, height variation, and dynamic balancing. 

\begin{figure}[h]
    \centering
    \includegraphics[width=\linewidth]{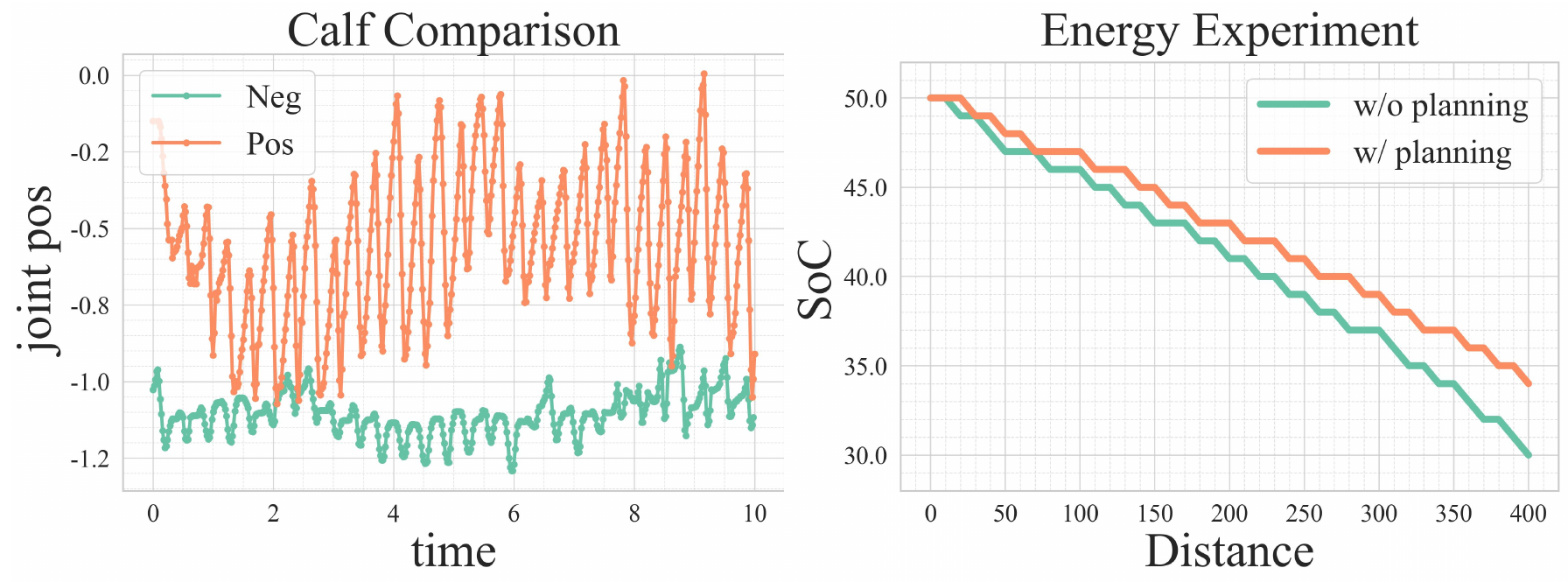}
    \vspace{-6.5mm}
    \caption{Left: \textbf{Calf position} under different planned joint targets. Right: \textbf{SoC} during the long-distance real-world evaluation.}
    \label{fig:energy_joint_stat}
    \vspace{-10pt}
\end{figure}

For energy saving task, we record the state of charge (SoC) of the robot every 10 meters. We initiate the robot to 50\% SoC, and command the robot to walk with 0.6 m/s. As shown in Fig.\ref{fig:energy_joint_stat}, the overall energy saving is 20\% with energy saving reward compared to no planning. The energy saving oriented agent demonstrate a smoother action pattern and decrease the foot lifting height to avoid unnecessary energy consumption that doesn't contribute to the actual movement. 

\begin{figure}[h]
    \centering
    \includegraphics[width=\linewidth]{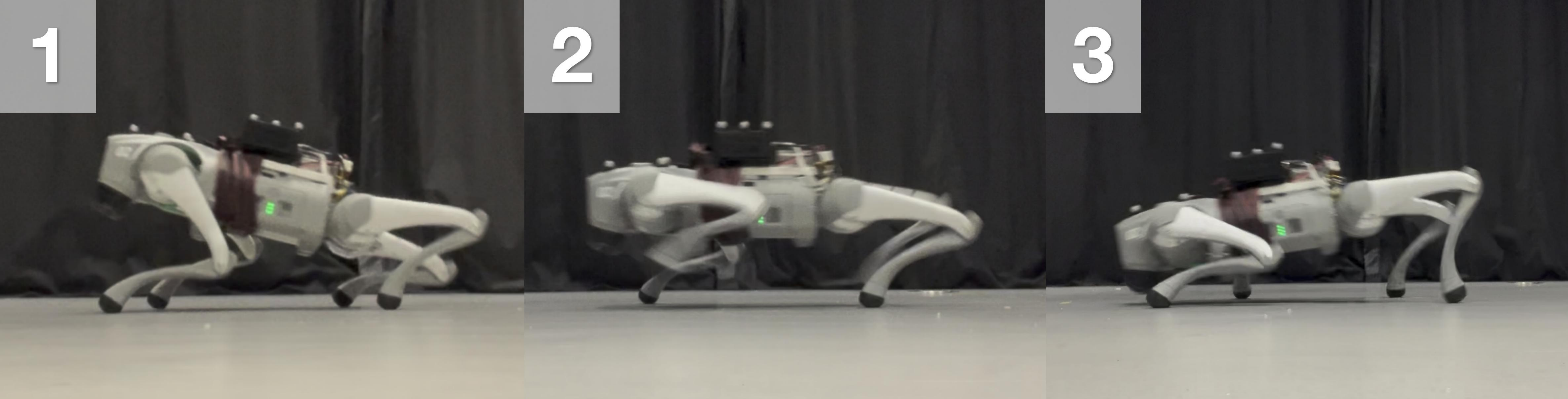}
    \vspace{-15pt}
    \caption{Comparison of behavior patterns with \textbf{different joint position targets}. From left to right: negative calf target, original planner, and positive calf target.}
    \label{fig:calf_comp}
    \vspace{-12pt}
\end{figure}

For joint position adaptation, two distinct reference joint poses are provided for the robot to track, which differ from the patterns of the source RL policy. The \textit{negative} setting enforces a relatively small calf angle (close to extension), while the \textit{positive} setting requires a larger angle (more flexed). Figure~\ref{fig:energy_joint_stat} shows the resulting rear-left calf trajectories. Under the positive target, the calf maintains a higher average position with large oscillations, indicating active participation in locomotion. In contrast, under the negative target the calf remains closer to extension with small oscillations, contributing less to propulsion. Figure~\ref{fig:calf_comp} further illustrates the resulting strategies: the negative target induces an upward tilt with propulsion dominated by the front legs, while the positive target shifts the robot forward, with larger calf–thigh angles and the front calves primarily used for balance.

\begin{figure}[t]
    \vspace{3.0mm}
    \centering
    \includegraphics[width=\linewidth]{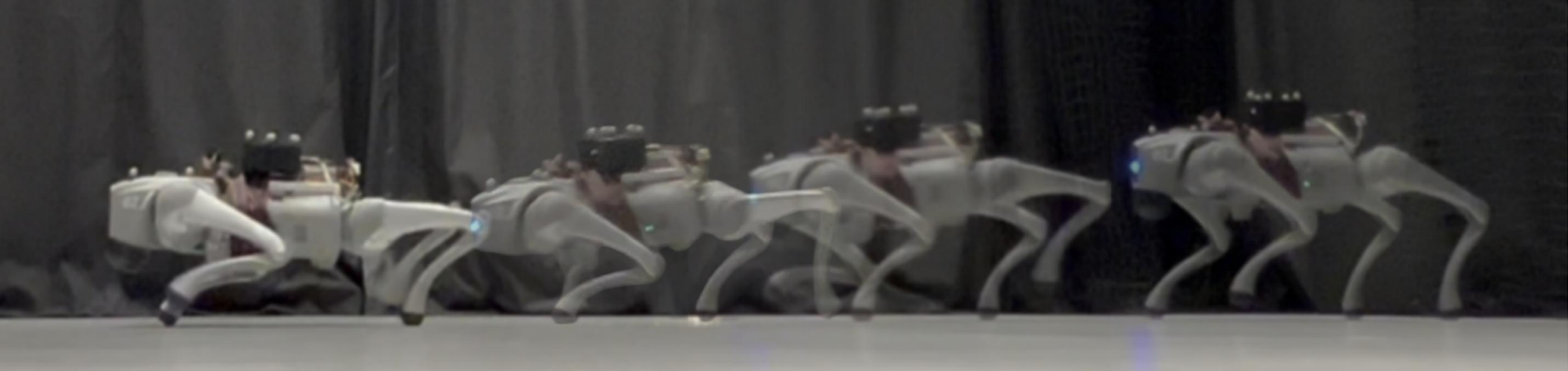}
    \vspace{-15pt}
    \caption{\textbf{Height transition} experiment.}
    \label{fig:height_pic}
    \vspace{-20pt}
\end{figure}

For height variation, a neural-network-based height reward model directs the robot from a relatively elevated base position to a lower one. As shown in Fig.~\ref{fig:height_pic}, the robot transitions from an initial height of 25\,cm to a reduced height of 18\,cm under reward planning. It is important to note that while the pretraining dataset includes demonstrations at both heights, it does not contain any demonstrations of transitions between them; the observed adaptation is therefore entirely achieved through test-time planning.
\begin{figure}[h]
    \centering
    \vspace{-2.5mm}
    \includegraphics[width=\linewidth]{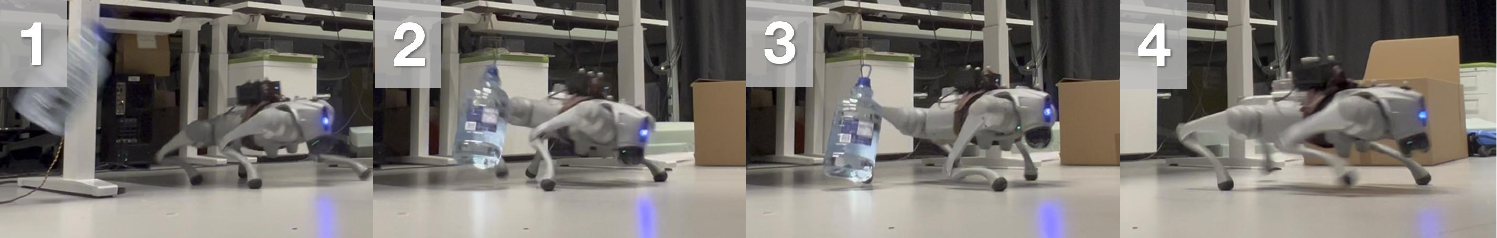}
    \vspace{-4.5mm}
    \caption{\textbf{Balancing under external disturbance.} The robot is subjected to a lateral push at the trunk and subsequently recovers from the perturbed posture, reestablishing balance.}
    \label{fig:balancing_pic}
    \vspace{-10pt}
\end{figure}

\begin{table}[htbp]
    \centering
    \vspace{-3mm}
    \setlength\tabcolsep{8pt}
    \begin{tabular}{lccc}
        \toprule
         & \textbf{Easy} & \textbf{Medium} & \textbf{Hard} \\
        \midrule
        \textbf{No Plan} & 0.9 & 0.8 & 0.6 \\
        \textbf{Plan}   & 1.0 & 1.0 & 0.9 \\
        \bottomrule
    \end{tabular}
    \caption{\textbf{Balancing} success rate under different pendulum angles.}
    \label{tab:balancing_stat}
    \vspace{-10pt}
\end{table}

For balancing, a pendulum apparatus is employed to generate controlled lateral impacts, similar to~\cite{xiao2024pa}. As shown in Fig.~\ref{fig:balancing_pic}, a 3.8\,kg weight suspended from a 1.2\,m pivot swings to strike the robot’s trunk at its lowest point, producing impact velocities of 4.54\,m/s (90° release) and 3.21\,m/s (60° release), corresponding to two difficulty levels. A trial is considered a failure if the robot falls or remains stuck below 0.2\,m/s for more than 2\,s. As summarized in Table~\ref{tab:balancing_stat}, incorporating balancing rewards and constraint planning markedly improves the ability to withstand external disturbances and aids recovery.

\subsubsection{Locomotion Performance}
\label{subsec:locomotion_result}
\begin{figure}[t]
    \vspace{3.0mm}
    \centering
    \begin{subfigure}{0.48\linewidth}
        \centering
        \includegraphics[width=\linewidth]{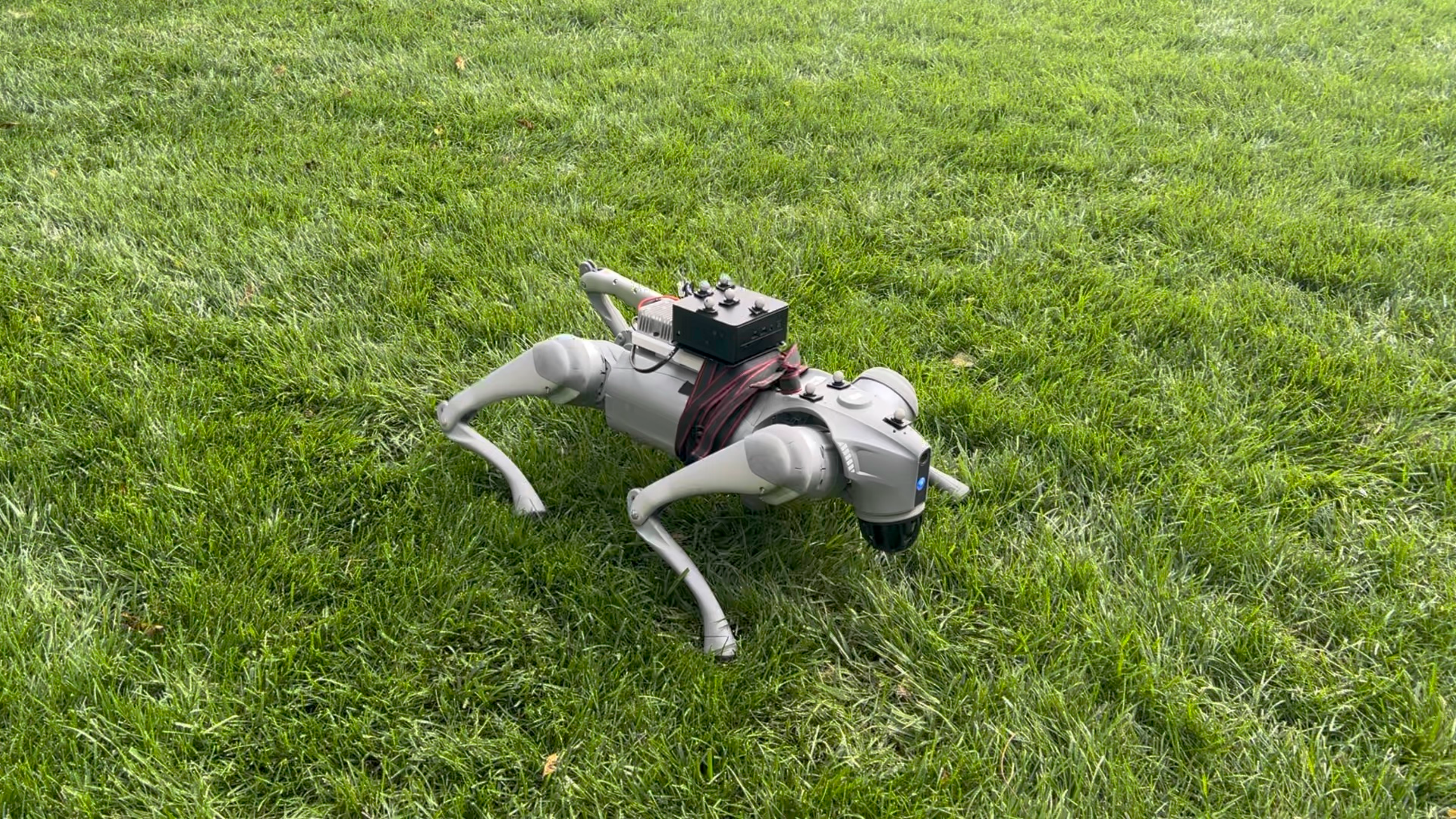}
        \label{fig:grass}
    \end{subfigure}
    \hfill
    \begin{subfigure}{0.48\linewidth}
        \centering
        \includegraphics[width=\linewidth]{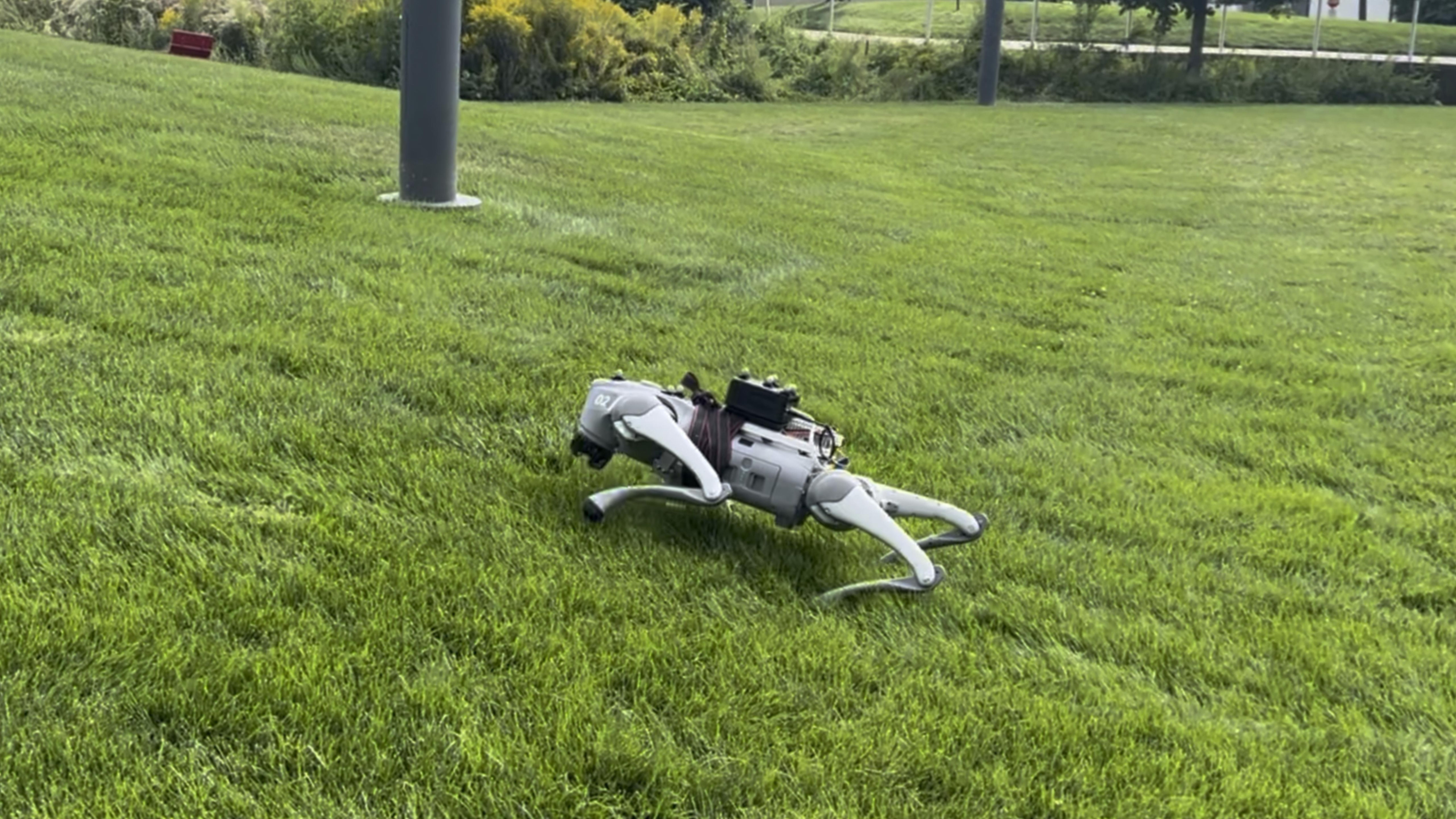}
        \label{fig:grassy_slope}
    \end{subfigure}
    \vspace{-3.5mm}
    \caption{\textbf{Zero-shot walking.} Left: grass. Right: grassy slope.}
    \label{fig:grass_combined}
    \vspace{-20pt}
\end{figure}

Diffusion-MPC is evaluated on challenging real-world terrains, including soft uneven grass with varying friction and a grass slope with varying inclination, as shown in Fig. \ref{fig:grass_combined}. The planner is deployed in a zero-shot manner without environment-specific retraining. A neural-network-based foot-lifting reward model encourages stable stepping on uneven surfaces, while a balancing reward enhances stability during traversal. For slope locomotion, regularization on the rear-calf joint position is applied adaptively: larger angles are favored for ascending slopes to prevent backward slipping, whereas smaller angles are encouraged for descending slopes to maintain forward stability. These results highlight that diffusion-based planning enables deployment in the wild, providing both adaptability to diverse terrains and flexible behavior modulation at test time.

\subsubsection{Interactive Learning Experiments}
\label{subsec:online_learning_result}
\begin{figure}[h]
    \centering 
    \vspace{-3mm}
    \includegraphics[width=\linewidth]{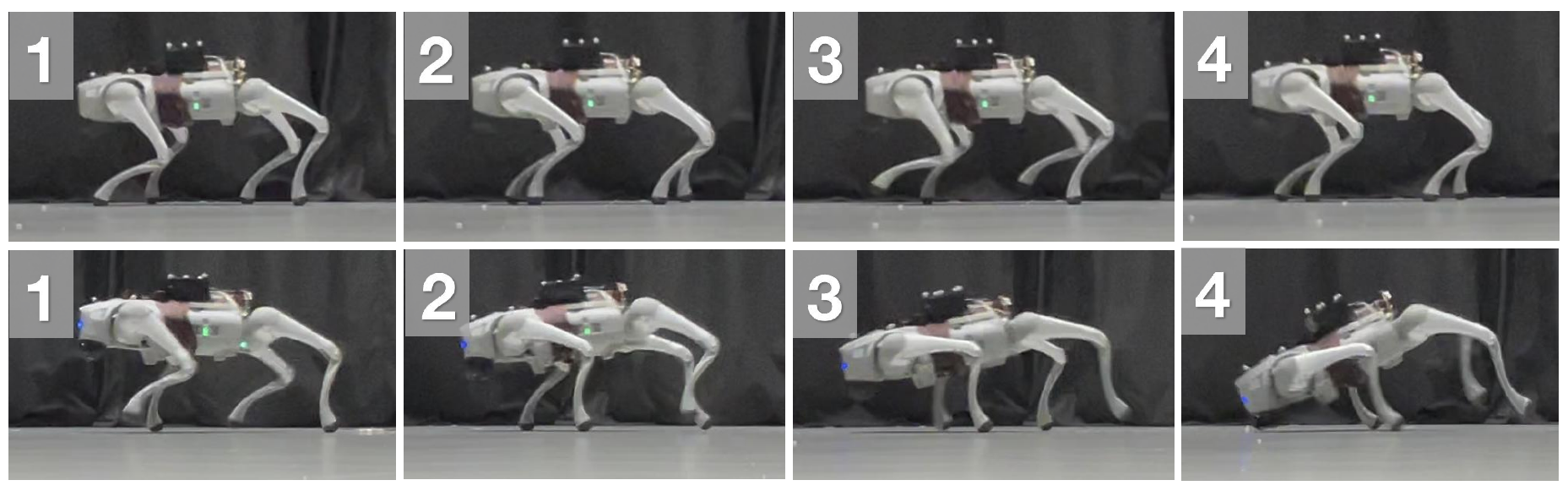}
    \vspace{-5mm}
    \caption{\textbf{Comparison of finetuned and untuned diffusion planners in real-world deployment.} 
The finetuned planner (top) achieves stable locomotion, while the untuned baseline (bottom) fails to deploy.}
    \label{fig:finetune-real}
    \vspace{-5pt}
\end{figure}

\begin{figure}[h]
    \centering 
    \vspace{-5.5mm}
    \includegraphics[width=\linewidth]{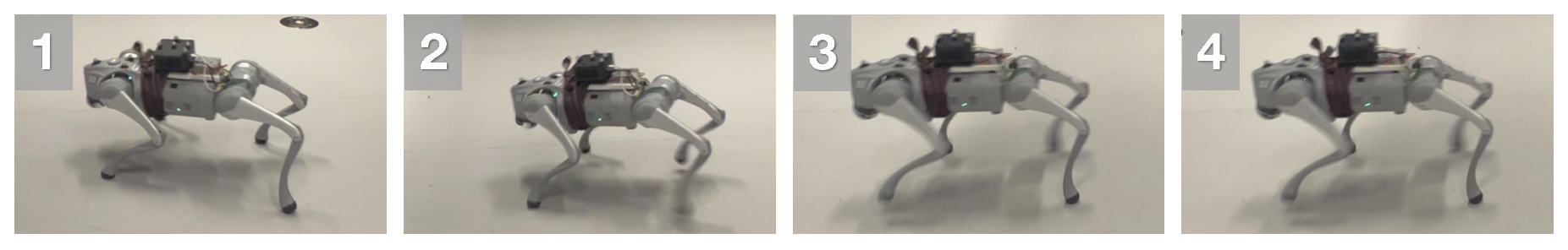}
    \vspace{-5mm}
    \caption{\textbf{Interactive diffusion planner learned from scratch} deployed into real world}
    \label{fig:scratch-real}
    \vspace{-15pt}
\end{figure}

As shown in Fig.~\ref{fig:finetune-real}, both the planner trained on datasets without domain randomization and its finetuned counterpart are evaluated in real-world experiments. As shown in Table~\ref{tab:finetune_real}, the robot is tested at different target speeds over a 10\,m trajectory, and success rates are reported. The finetuned policy exhibits substantial gains over the untuned version. By contrast, the untuned policy performs even worse than in simulation, likely due to the replan margin during deployment, which reduces responsiveness. In addition, Fig.~\ref{fig:scratch-real} presents real-world deployment of a planner trained entirely from scratch, demonstrating that diffusion-based planning can be learned effectively without reliance on demonstration data.

\begin{table}[htbp]
    \vspace{-2mm}
    \centering
    \setlength\tabcolsep{8pt}
    \fontsize{8.5}{11}\selectfont
    \begin{tabular}{l|cc}
        \toprule
        \textbf{Goal (Task)} & \textbf{Finetuned} & \textbf{Not Tuned} \\
        \midrule
        \textbf{0.5 m/s} & \textbf{1.0} & 0.2  \\
        \textbf{0.7 m/s} & \textbf{1.0} & 0.1  \\
        \textbf{1.0 m/s} & \textbf{1.0} & 0.0   \\
        \bottomrule
    \end{tabular}
    \caption{\textbf{Effectiveness of Finetuning.} 
Success rates (percentage of successful runs) are reported at different commanded forward speeds.}
    \label{tab:finetune_real}
    \vspace{-10pt}
\end{table}

\subsubsection{Deployment Ablation}
We conduct and ablation study on how the choice of replan margin $D$, caching steps $m$, and cache reset influence frequency, latency, and real-world locomotion performance. Performance is assessed along three criteria: forward, turning, and backward locomotion. Since aggressive caching may affect the transition between locomotion patterns, sequential omni-directional movement is specifically tested. The robot is initialized in a standing position, then commanded to move forward at $V_x = 0.8$\,m/s for 3\,s, followed by combined motion at $V_x = 0.5$\,m/s and $V_{\text{yaw}} = 1.0$\,m/s, and finally commanded to walk backward at $V_x = -0.8$\,m/s. Latency is recorded once the computation reaches steady state, excluding rare deviations caused by resets. Results in Table~\ref{tab:deployment_ablation} show that DDIM degrades motion quality compared to cached diffusion steps, and that cache resetting is crucial for robust omni-directional transitions. With these designs, diffusion planning achieves real-time operation in onboard computer while preserving motion quality.

\begin{table}[htbp]
    \centering
    \vspace{2mm}
    \setlength\tabcolsep{4pt}       
    \renewcommand{\arraystretch}{1.05} 
    \fontsize{8}{10}\selectfont
    \begin{tabular}{p{0.16\linewidth}|ccccc}
        \toprule
        \textbf{Metrics} & \textbf{A} & \textbf{B} & \textbf{C} & \textbf{D} & \textbf{E} \\
        \midrule
        Frequency        & 33.1 & 45.8 & 47.7 & 50.1 & 50.0 \\
        Latency          & 198  & 77   & 78   & 22   & 21   \\
        Forward          & \xmark & \xmark & \xmark & \cmark & \cmark \\
        Turning          & \xmark & \xmark & \xmark & \cmark & \cmark \\
        Backward         & \xmark & \xmark & \xmark & \xmark & \cmark \\
        \bottomrule
    \end{tabular}

    \vspace{2pt}
    {\footnotesize
    \textbf{NOTE:}
    A: $D{=}0$, $m{=}0$;\;
    B: $D{=}0$, $m{=}7$, \emph{no refresh};\;
    C: \emph{DDIM (3 steps)};\;
    D: $D{=}3$, $m{=}7$, \emph{no refresh};\;
    E: $D{=}3$, $m{=}7$, \emph{refresh}.
    }
    \caption{\textbf{Deployment Ablation.}
    Impact of replan margin $D$, caching steps $m$, and cache refresh. We report frequency (hz), latency (ms), and real-world locomotion performance across configurations.}
    \label{tab:deployment_ablation}
    \vspace{-18pt}
\end{table}
\section{CONCLUSION}
We present Diffusion-MPC, a planning framework that leverages diffusion models as expressive generative priors over state–action trajectories, synthesizing generalized behaviors driven by a broad range of rewards and constraints. Diffusion-MPC supports novel reward functions and constraints at test time, allowing flexible adaptation to new tasks and environments. Our approach leverages the dataset to capture and compose existing behaviors into novel ones, and further employs an online finetuning mechanism that actively explores the task space to recover behaviors absent from demonstrations, enhancing overall performance. Diffusion-MPC runs in real time via asynchronous planning and early-step caching, and exhibits flexible behavior adaptation. Our work positions diffusion-based generative priors as a practical path to adaptable, general-purpose embodied control.

Future directions include extending the framework to integrate diverse inputs—such as LiDAR, visual perception, and natural language—advancing toward embodied agents capable of fully autonomous real-world interaction and planning. Our present study considers only simple constraint classes; extending the approach to richer, task-dependent (and potentially nonconvex or time-varying) constraints is a natural next step. Another key direction is to develop interactive finetuning methods that can train competitive planners without relying on offline data, thereby removing the need for expert demonstrations while preserving sample efficiency.

\vspace{-5pt}
\bibliographystyle{./IEEEtran} 
\bibliography{./IEEEabrv,./IEEEexample}



\clearpage
\newpage


\end{document}